# Video-Based Performance Evaluation for ECR Drills in Synthetic Training Environments

**Surya Rayala, Marcos Quinones-Grueiro, Naveeduddin Mohammed, Ashwin T S, Gautam Biswas**

**Vanderbilt University**

Nashville, TN

surya.chand.rayala@vanderbilt.edu,
marcos.quinones.grueiro@vanderbilt.edu,
naveeduddin.mohammed@vanderbilt.edu,
ashwin.tudur.sadashiva@vanderbilt.edu,
gautam.biswas@vanderbilt.edu

**Benjamin Goldberg, Randall Spain, Paige Lawton**

**US Army DEVCOM Soldier Center - STTC**

Orlando, FL

benjamin.s.goldberg.civ@army.mil,
randall.d.spain.civ@army.mil
paige.m.lawton.civ@army.mil

## ABSTRACT

Effective urban warfare training requires situational awareness and muscle memory, developed through repeated practice in realistic yet controlled environments. A key drill, Enter and Clear the Room (ECR), demands threat assessment, coordination, and securing confined spaces. The military uses Synthetic Training Environments (STEs) that offer scalable, controlled settings for repeated exercises. However, automatic performance assessment remains challenging, particularly when aiming for objective evaluation of cognitive, psychomotor, and teamwork skills. Traditional methods often rely on costly, intrusive sensors or subjective human observation, limiting scalability and accuracy. This paper introduces a video-based assessment pipeline that derives performance analytics from training videos without requiring additional hardware. By utilizing computer vision models, the system extracts 2D skeletons, gaze vectors, and movement trajectories. From these data, we develop task-specific metrics that measure psychomotor fluency, situational awareness, and team coordination. These metrics feed into an extended Cognitive Task Analysis (CTA) hierarchy, which employs a weighted combination to generate overall performance scores for teamwork and cognition. We demonstrate the approach with a case study of real-world ECR drills, providing actionable, domain-specific metrics that capture individual and team performance. We also discuss how these insights can support After Action Reviews (AARs) with interactive dashboards within Gamemaster and the Generalized Intelligent Framework for Tutoring (GIFT), providing intuitive and understandable feedback. We conclude by addressing limitations, including tracking difficulties, ground-truth validation, and the broader applicability of our approach. Future work includes expanding analysis to 3D video data and leveraging video analysis to enable scalable evaluation within STEs.

## ABOUT THE AUTHORS

**Surya Rayala** is a Ph.D. student in Computer Science at Vanderbilt University. His research focuses on computer vision, primarily leveraging and developing methods for multi-object detection, tracking, action recognition, and person re-identification in challenging simulation environments, advancing automated performance assessment for training scenarios such as army battle drills and nursing simulations.

**Dr. Marcos Quinones-Grueiro** is a research scientist at the Institute for Software Integrated Systems and an adjunct professor in the Computer Science department at Vanderbilt University. His research interests are developing and applying foundational machine learning and reinforcement learning methods across crucial areas such as smart mobility, multimodal analytics in education, safety in aerospace operations, and energy management optimization. He has over 100 refereed publications, and his research has been supported by funding from the Army, NASA, and NSF.

**Naveeduddin Mohammed** is a Lead Engineer with the Institute for Software Integrated Systems at Vanderbilt University. Naveed received his M.S. degree in Computer and Information Sciences from University of Colorado. He is a full stack developer specializing in the design, planning, implementation, deployment, testing and maintenance of frameworks for open-ended computer-based learning environments and metacognitive tutors. His work focuses on developing and enhancing computer-based learning environments, particularly in the areas of teachable agents, learning-by-teaching and learning-by-modeling. He is involved in creating and integrating novel computational paradigms to advance simulation-based and multimodal learning systems.

**Dr. Ashwin T S** is a Research Scientist at the Institute for Software Integrated Systems, Computer Science, Vanderbilt University, Nashville, Tennessee. His research focuses on Computer Vision, Multimodal Machine Learning, Affective Computing, Learning Technologies, and Human-Computer Interaction. Dr. Ashwin's work has significantly contributed to video affective content analysis, multi-modal learning analytics, and the application of deep learning in education.

**Dr. Gautam Biswas** is the Cornelius Vanderbilt Professor of Engineering and Professor of Computer Science and Computer Engineering at Vanderbilt University. He researches Intelligent Systems with a focus on creating adaptive open-ended learning environments that respond to students' learning performance and behaviors. He has created innovative multimodal analytics for analyzing students' learning behaviors in various simulation and augmented reality-based training environments. He has authored over 700 refereed publications. He is a Life Fellow of the IEEE, a Fellow of the Asia Pacific Society for Computers in Education, and a Fellow of the Prognostics and Health Management Society.

**Dr. Benjamin Goldberg** is a Senior Research Scientist at the U.S. Army Combat Capability Development Command (DEVCOM) - Soldier Center. He serves as technical lead for the Adaptive & Intelligent Training Systems branch, where their research focuses on adaptive experiential learning with an emphasis on simulation-based environments and leveraging Artificial Intelligence to create personalized experiences. Dr. Goldberg holds a Ph.D. in Modeling & Simulation from the University of Central Florida and is well-published across several high-impact journals and proceedings, including IEEE Transactions on Learning Technologies, the Journal of Artificial Intelligence in Education, and Computers in Human Behavior.

**Dr. Randall Spain** is a Research Scientist at the U.S. Army Combat Capability Development Command – Soldier Center. He holds a Ph.D. in Human Factors Psychology from Old Dominion University and an M.S. in Experimental Psychology. His research focuses on designing and evaluating adaptive and intelligent training systems.

**Dr. Paige Lawton** is a Research Psychologist at the U.S. Army Combat Capability Development Command – Soldier Center. She holds a Ph.D. in Human Factors Psychology from Embry-Riddle Aeronautical University. Her research focuses on developing and evaluating adaptive and intelligent training systems.

# Video-Based Performance Evaluation for ECR Drills in Synthetic Training Environments

**Surya Rayala, Marcos Quinones-Grueiro, Naveeduddin Mohammed, Ashwin T S, Gautam Biswas**

**Vanderbilt University**

**Nashville, TN**

surya.chand.rayala@vanderbilt.edu,
marcos.quinones.grueiro@vanderbilt.edu,
naveeduddin.mohammed@vanderbilt.edu,
ashwin.tudur.sadashiva@vanderbilt.edu,
gautam.biswas@vanderbilt.edu

**Benjamin Goldberg, Randall Spain, Paige Lawton**

**US Army DEVCOM Soldier Center - STTC**

Orlando, FL

benjamin.s.goldberg.civ@army.mil,
randall.d.spain.civ@army.mil
paige.m.lawton.civ@army.mil

## 1. INTRODUCTION

Repeated practice and objective assessment are crucial for maintaining psychomotor skills, situational awareness, and muscle memory in complex military training (Tosuncuoglu, 2018; Vatral et al., 2021). Traditionally, scenarios such as Enter and Clear a Room (ECR) have relied on expert observation to support After-Action Reviews (AARs). While expert evaluations are valuable, they are costly, subjective, and often miss subtle performance details across sessions (Goldberg et al., 2020; Vatral et al., 2022a). Accordingly, interest has grown in supplementing expert feedback with data-driven insights. Sensor-based systems—including inertial measurement units (IMUs), motion capture (Mishra et al., 2021), and eye trackers (Yang et al., 2024)—extract behavioral signals such as skeletal pose, gaze, and spatial position and map them to cognitive, psychomotor, and affective skills (Blikstein & Worsley, 2016). However, these systems are expensive, require complex setup and calibration, demand ongoing maintenance and trained personnel, and can be intrusive, disrupting natural performance. Conversely, strategically placed video cameras remain underused for automated assessment. Advances in real-time computer vision, particularly methods that extract two-dimensional body keypoints from image frames, now enable the collection of detailed behavioral data from standard video (Munea et al., 2020; Chen et al., 2022).

This paper presents a video-based pipeline that extends keypoint models to extract 2D skeletons, estimate gaze direction, and infer spatial positions of participants in a battle drill. These outputs automate the evaluation of soldier behavior and performance in Synthetic Training Environments (STEs). To demonstrate effectiveness, we conduct a case study on the ECR drill (Section 2), analyzing the video to compute actionable metrics. These metrics are integrated into a Cognitive Task Analysis (CTA) hierarchy—developed for ECR—to assess higher-level cognitive and teamwork constructs (Vatral et al., 2022a).

The remainder of the paper is organized as follows. We describe the case study setup, detailing the ECR drill within the Small Unit Performance Analytics (SUPRA) environment and the video characteristics used for analysis. We then present the CTA framework, including the hierarchical model and the integration of new metrics. Next, we outline the video analysis pipeline, metric definitions and derivations, and hierarchical performance aggregation. We report performance results across multiple teams and trials, with key observations. Finally, we discuss implications, integration with AAR systems, limitations, opportunities for future research, summarize contributions, and conclude the paper.

## 2. CASE STUDY

To evaluate the effectiveness of our video analysis pipeline and derived performance metrics, we conducted a case study using real-world data from live Enter and Clear a Room (ECR) drills within SUPRA training exercises. ECR (Battle Drill 6) tasks a squad with rapidly entering a room, securing the entry point, neutralizing threats, and ensuring the space is safe before exiting. From a doctrinal point of view, soldiers enter the room in sequence, move along walls to avoid crossfire and blind spots, establish points of dominance, neutralize threats within their sectors of fire, and systematically clear the room before issuing the all-clear.

The dataset comprises video recordings from 19 squads operating in a live training environment with real equipment and personnel. Each squad executed three trials of the same ECR scenario across three consecutive days. The facility was a squad house with three rooms and a hallway; each room presented a distinct challenge: tactical targets, surrendering combatants, or armed enemy combatants. Videos were recorded from multiple vantage points. Due to inconsistencies in quality—occlusions, low lighting, and dark uniforms blending with the background—we focused on Room 2 (armed enemy combatants), where the primary objective is threat neutralization and room security. This subset comprises 54 videos, one for each trial, representing 18 squads with team sizes ranging from two to four. For this study, we analyzed videos from four squads sampled over a three-year period to test the robustness of our approach under varying lighting conditions and camera angles.

## 3. CTA FRAMEWORK

CTA is a well-established method for breaking down complex tasks into their cognitive, behavioral, and team-related components (Clark & Estes, 1996; Zachary et al., 2000). In military training and evaluation, CTA models are especially useful because they reveal the layered structure of expertise, connecting observable trainee actions to underlying cognitive, psychomotor, and teamwork constructs (Clark & Estes, 1996). Typically, CTA models are developed through an iterative process involving doctrinal review, expert interviews, and systematic observation of the task. In this work, we focus on modifying and applying a previously developed CTA for ECR scenarios, where the task is inherently collaborative and team performance takes priority over individual evaluation, as seen in prior studies (Vatral et al., 2022a). Our updated CTA model is shown in Figure 1.

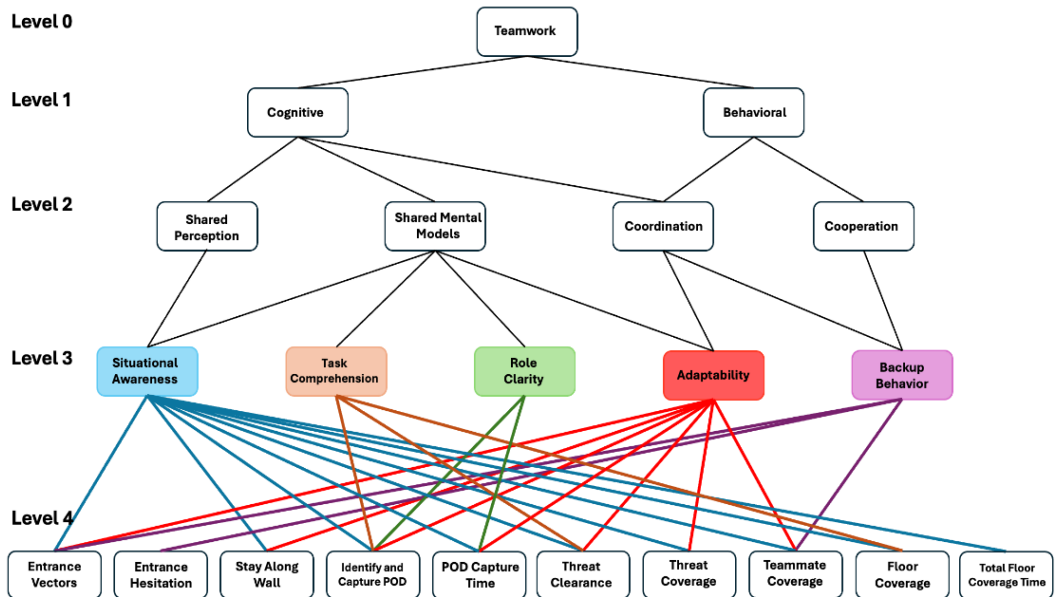

**Figure 1. The cognitive task model developed for this study**

We maintained the structure and definitions of Levels 0 to 3 from our previous version of the model (Vatral et al., 2022a). The main improvements to the CTA are at Level 4, where we introduce and enhance a set of ten performance metrics that are directly calculated by our video analysis pipeline. Three of these metrics —Entrance Vectors, Entrance Hesitation, and Stay Along Wall —were adapted from earlier work, incorporating improved computational methods. The other seven metrics are newly introduced in this study and leverage additional features extracted from skeletal and gaze data, made possible by our video analytics pipeline. The development and selection of these metrics involved close collaboration with subject matter experts and an extensive review of doctrinal sources. After creating the new metrics, we identified the appropriate linkages from Level 4 to Level 3 constructs.

While our current model relies on video-derived metrics, future versions can incorporate additional metrics and modalities, such as physiological signals and audio data. The connections between metrics and higher-level constructs are based on the definitions and intended purposes of each metric. For example, the Level 3 construct Task Comprehension is informed by three metrics: Identify and Capture POD, Threat Clearance, and Floor Coverage. These metrics directly indicate whether the team achieved the core goal of the ECR Task—clearing the room of threats. Other metrics, though not directly tied to mission completion, assess the quality of execution. For instance, Stay Along the Wall relates to Situational Awareness and Adaptability because it shows the team's ability to avoid blind spots and navigate different room layouts efficiently. Detailed definitions and calculation methods for all ten metrics are provided in Section 4.2.

## 4. METHODS

Our video analysis and metrics derivation pipeline is divided into three distinct phases. The first phase focuses on detecting and extracting skeletal, positional, and gaze information from participant video recordings. In the second phase, the extracted data are processed to generate domain-specific metrics relevant to the study's objectives. Finally, the third phase involves aggregating these metric values to higher levels within the CTA framework, as described in Section 3.

### 4.1. Extraction of Skeleton, Position, and Gaze Information from Video

The four primary steps that define our video analysis are described below.

#### 4.1.1. Extracting Bounding Boxes and Skeletal Points from Each Video Frame

We identify and localize participants in each video frame using a two-stage, top-down approach leveraging recent advances in real-time human pose estimation (Chen et al., 2022). First, the RTMDet-m detector locates individuals and produces bounding boxes (Lyu et al., 2022). These boxes and the corresponding RGB frames are then passed to RTMPose-x to estimate skeletal keypoints (Jiang et al., 2023). This two-stage design combines the strengths of both architectures, yielding speed and accuracy with a moderate number of people per frame (Jiang et al., 2023). Figure 2 illustrates this process.

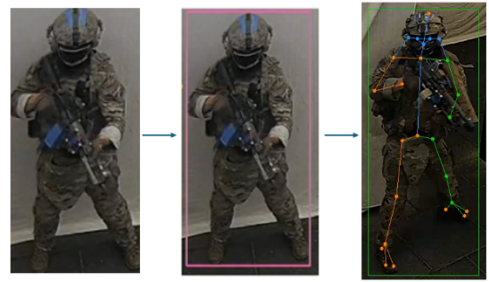

**Figure 2. Two Stage Detection**

Due to the limited availability of task-specific data, we utilize pre-trained models for detection and pose estimation, predicting the Halpe26 keypoint set and adhering to AlphaPose protocols for comprehensive coverage of human keypoints (Fang et al., 2023). These models were initially trained on seven large-scale public datasets and demonstrate robust generalization (MMPose Model Zoo, 2025).

However, we observed a performance gap in detection accuracy for soldier-specific appearances, as public datasets contain little representation of soldiers. Building on the findings that improved detection directly enhances downstream pose estimation accuracy (Jiang et al., 2023), we further fine-tuned the RTMDet model using a custom dataset. We manually annotated bounding boxes for frames randomly sampled from videos of all 19 SUPRA squads, totaling 3,128 training frames and 374 validation frames. Figure 3 shows results indicating the increase in mean Average Precision (mAP; higher is better) achieved by fine-tuning RTMDet on our dataset.

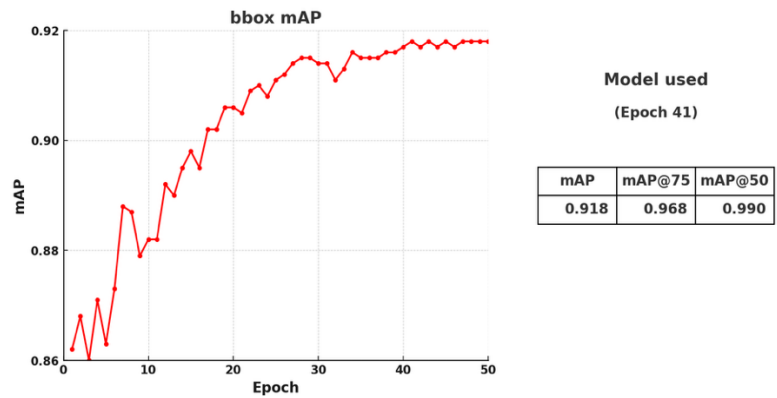

**Figure 3. Fine-Tuning RTMDet**

### 4.1.2. Tracking Detections Across Continuous Video Frames

After extracting bounding boxes and skeletal keypoints for all individuals in each frame, it is important to keep consistent identities for each person across frames. This enables the temporal linking of detections from Section 4.1.1 for each person through the entire video sequence. Since uniformed soldiers look very similar and often occlude each other in the small SUPRA training area, advanced Deep Learning-based appearance tracking methods (Park et al., 2021) were ineffective. Instead, we used OC-SORT, a fast and reliable tracking algorithm that predicts each person's location using a Kalman filter and matches detections based on motion. OC-SORT also corrects errors that may occur during occlusions by utilizing the latest observations before and after an object is temporarily lost and rediscovered (Cao et al., 2023). We further optimized OC-SORT parameters for our videos. Figure 4 illustrates how distinct IDs are assigned to different individuals.

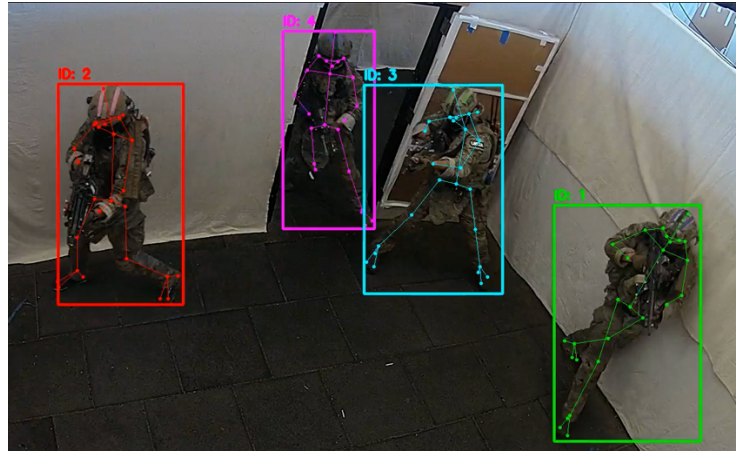

**Figure 4. Tracking Individuals with ID's**

### 4.1.3. Extracting Position Information into a Topological Room Map

For spatial analysis of participant movement within the exercise space, we build upon the planar projection approach proposed by (Vatral et al., 2021), where 2-D video positions are mapped onto a top-down topological map of the room. Our method extends this work by incorporating confidence-based keypoint selection, Kalman-filter-based smoothing, and dynamic adjustment for missing data.

For each frame, we first identify the most reliable keypoints for each individual, as recorded using the methods discussed in Sections 4.1.1 and 4.1.2. When high-confidence ankle, heel, or toe keypoints are available at the time $t$, the position with respect to the video coordinate system $P_t$ is computed as the average of these points, as shown in equation 1, where $(x_i, y_i)$ are the coordinates of foot keypoints on the video frame and $K$ is the aforementioned set of valid keypoints and $N = |K|$.

$$P_t = \frac{1}{N} \sum_{i \in K} \begin{bmatrix} x_i \\ y_i \end{bmatrix} \quad (1) \qquad v = \frac{s_t - s_{t-k}}{k} \quad (2) \qquad P_{t+1} = P_t + v \quad (3)$$

To smooth the estimated movement paths and handle short-term detection failures, we utilize a two-dimensional Kalman filter that is updated with each valid foot position. Suppose key foot keypoints are missing for several frames. In that case, we estimate velocity by examining the history of other available key points or the bounding box positions and interpolating the previously available foot point-based position. Specifically, the velocity $v$ is calculated as in equation 2, where $s_t$ and $s_{t-k}$ are the locations of the other available keypoints (or bounding box centers) at frames $t$ and $t-k$, respectively. This velocity estimate is then used to predict the next position, $P_{t+1}$, when keypoints are temporarily missing (see equation 3), where $P_t$ is the last known position and $v$ is the estimated velocity calculated as above. This helps maintain a smooth and continuous trajectory, even during brief periods of detection failure.

$$\begin{bmatrix} x_m \\ y_m \\ 1 \end{bmatrix} = H \begin{bmatrix} x_p \\ y_p \\ 1 \end{bmatrix} \quad (4) \qquad M_t^{\text{smooth}} = \alpha\, M_t + (1 - \alpha)\, M_{t-1}^{\text{smooth}} \quad (5)$$

The foot position in the image is then projected onto the room map using a planar homography, as shown in equation 4. Here, $(x_p, y_p)$ is the pixel-space position point, and $(x_m, y_m)$ is its top-down map coordinate, $H$ is the 3 × 3 planar-homography matrix built from selecting four or more well-separated reference points that are easily identifiable in both the video frame and the room map (Vincent et al., 2001). This allows us to accurately align pixel positions from the video to the map. Finally, to minimize jitter in the mapped trajectory points, we apply dynamic smoothing by blending the currently observed $M_t$ and previous smoothed map positions $M_{t-1}^{smooth}$, as shown in equation 5, where the smoothing factor $\alpha$ is chosen based on the magnitude of movement between frames.

Figure 5 illustrates the mapping process. Each team member receives a unique ID number upon entering the room through the designated entry area on the map. Armed personnel already inside the room who do not enter through the door are identified as enemy combatants and marked in white. While enemy positions are tracked, only the trajectories of team members are shown in the figure, keeping the focus on the squad's movements.

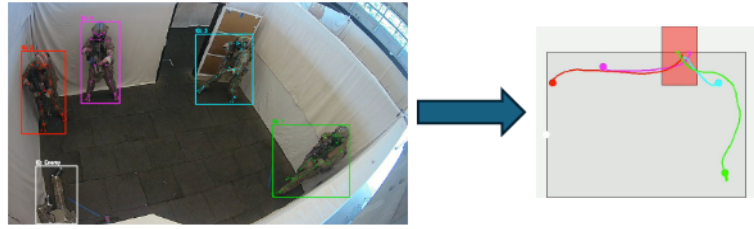

**Figure 5. Position Mapping**

### 4.1.4. Extracting Gaze Focus Triangle

To estimate each participant's gaze direction from the video, we compute a 2-D gaze vector using keypoints on the head recorded using methods described in Sections 4.1.1 and 4.1.2. The gaze origin is computed as the midpoint between the two eye locations, when available. If only one eye is confidently detected, that eye's position is used. If eye keypoints are unreliable, the nose location is used as a fallback. The gaze direction $g$ is calculated as the vector from the midpoint between both ears to the center of the eyes or nose points detected. This direction vector is normalized to unit length as shown in equation 6, where $o$ is the gaze origin and $e$ is the ear midpoint.

$$\textbf{\textit{Gaze Direction}}: g = \frac{o-e}{||o-e||} \quad (6)$$

$$v_0 = origin \quad (7)$$
$$v_1 = origin + R(+\theta) \cdot g \cdot L \quad (8)$$
$$v_2 = origin + R(-\theta) \cdot g \cdot L \quad (9)$$

Identifying the exact gaze target directly is not possible in 2-D video, so we approximate the participant's gaze region by constructing an isosceles "gaze focus triangle." The triangle has its vertex at the gaze origin, extends in the gaze direction, and spans a visual angle parameter which is set to 20° (±10° around the vector). The vertices $v_n$ are computed as in equations 7,8,9 where $R(\pm\theta)$ are 2D rotation matrices for ±10°, and L is the triangle's length *(In our experiments, the triangle length L is extended until the vector intersects with a wall or the boundary of the room, since we assume there is no fixed limit to a participant's line of sight indoors).* An example of this is shown in Figure 6, This triangle with visual angle 20° represents the likely region of visual focus based on head orientation (Wang et al., 2019). For analysis, we check which objects or teammates are intersected by a participant's gaze triangle, indicating their attention focus in the scenario.

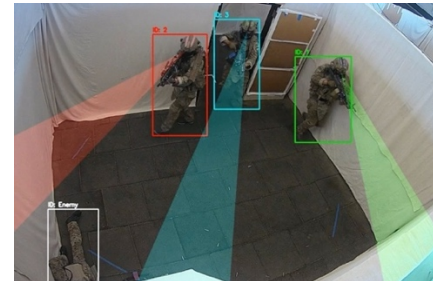

**Figure 6. Gaze Triangles**

### 4.2. Computation of Metrics

This section defines the metrics aligned with the CTA model (Section 3) and details their computation from video-derived signals (Section 4.1). Metrics are tailored to ECR in the SUPRA environment and were informed by prior research (Vatral et al., 2022a), Army doctrine (Department of the Army, 2007, 2017), instructional materials (Boise State University, 2014), and expert input.

1. **Entrance Vectors**. Assesses alternation of entry directions (e.g., left–right–left) using each soldier's entry position and post-door movement vector (Section 4.1.3). Score: proportion of consecutive entry pairs that alternate directions (range 0–1).
2. **Entrance Hesitation**. Evaluates the timeliness of entries via inter-entry gaps computed from mapped positions (Section 4.1.3). Uses two thresholds (general and the longer between the second and third entrants), both of which are instructor-adjustable. Score: average exponential penalty on entry delays (range 0–1).

3. **Identify and Capture POD**. Measures whether each soldier reaches and holds an assigned Point of Dominance (POD) using mapped positions (Section 4.1.3). PODs are auto-assigned from a predefined set based on first entry direction; instructors may redefine optimal Score: proportion of assigned PODs successfully captured (range 0–1).
4. **POD Capture Time**. Penalizes delays in reaching assigned PODs within instructor-set time limits. Full credit is given if within the limit; late arrivals are penalized; and missed POD yields zero. Score: average exponential penalty on POD capture delays (range 0-1).
5. **Move Along the Wall**. Verifies that each soldier remains within an instructor-set distance from walls from entry to POD (Sections 4.1.3 and metric 3). Score: average fraction of time soldiers remain inside the wall buffer before reaching POD (range 0–1).
6. **Threat Clearance**. Determines whether each enemy is neutralized using bounding boxes, keypoints (Sections 4.1.1–4.1.2), and gaze triangles (Section 4.1.4). For each enemy, it requires a minimum overlap duration with any soldier, plus wrist contact and direct gaze within that window; instructors set all thresholds. Score: proportion of detected enemies successfully cleared (range 0–1).
7. **Threat Coverage**. Measures visual coverage of uncleared enemies using bounding boxes, keypoints, and gaze (Sections 4.1.1–4.1.4) and the clearance status from metric 6. For frames with at least one trainee and one uncleared enemy, checks if each uncleared enemy is observed by at least one teammate (gaze–box intersection). Score: proportion of the time that uncleared enemies are kept in view by at least one teammate (range 0–1).
8. **Teammate Coverage**. Ensures each soldier is visible to at least one teammate using gaze–box intersections (Sections 4.1.1–4.1.2, 4.1.4). Score: 1 − (total time any soldier is unseen by all teammates)/(total team time in room) (range 0–1).
9. **Floor Coverage**. Projects each trainee's gaze triangle onto the floor via planar homography (Sections 4.1.3–4.1.4) and accumulates unique areas covered over time. Score: fraction of floor area visually covered by the team at the end of the run (range 0–1).
10. **Total Floor Coverage Time**. Checks if the entire floor is visually scanned within an instructor-set time limit. Score: average exponential penalty on team floor-coverage delay (range 0–1).

### 4.3. CTA Roll-Up

Multiple approaches can be used to aggregate performance metrics within the CTA hierarchy. Notably, Weighted Aggregation, as in (Vatral et al., 2022b), and Bayesian Roll-up, as in (Vatral et al., 2022a), have both been employed in the literature to combine low-level performance metrics into higher-level team and cognitive constructs. Weighted aggregations are computationally efficient and easy to extend, but do not account for uncertainty or latent state transitions. In contrast, Bayesian roll-ups (e.g., H-ABC Bayesian frameworks) offer a rigorous approach to modeling latent knowledge states and their changes across trials, albeit at the expense of exponential growth in Conditional Probability Tables (CPTs) as the hierarchy and metric set expand. A particular challenge in our scenario is the absence of CPTs, and the potential addition of future metrics makes comprehensive CPT design impractical. In the future, we will explore approximation methods, such as the Noisy-Or method, for establishing CPTs for our Bayes net representations, where all variables are discrete (Onisko et al., 2001). In our current approach, we have adopted a weighted aggregation method to combine low-level metrics into higher-order cognitive, teamwork, and psychomotor skill nodes.

#### 4.3.1. Weighted Hierarchical Aggregations with Exponential Smoothing Across Trials

To combine performance over multiple trials of the drill, we use a weighted hierarchical roll-up with exponential smoothing across trials. This approach builds on standard weighted propagation in the CTA hierarchy by adding time-dependent smoothing, enabling each node to gather and update its value based on both recent trial results and past performance history.

First, for each trial, before assignment of the performance score for that trial, a parent node's value $S_p^{(t)}$ for trial $t$ is computed as the weighted average of its children's values, as shown in equation 10, where $x_c^{(t)}$ is the value of the child node $c$ at trial $t$, $W_p$ is the sum of weights for all children $c$ of parent $p$ and $w_{c \to p}$ is the corresponding weight.

Next, at each node in the hierarchy, the final assigned performance score for trial $t$ is computed as an exponentially smoothed combination of the previous score and the current trial's value, as in equation 11, where $\widehat{s_p^{(t)}}$ is the smoothed (assigned) score for node $p$ at trial $t$, $\widehat{s_p^{(t-1)}}$ is the assigned score of the node score from the previous trial, $S_p^{(t)}$ is the current trial's node value computed for node $p$, and $\alpha(t)$ is a smoothing factor.

$$S_p^{(t)} = \frac{1}{W_p} \sum_c w_{c \to p}\, x_c^{(t)} \quad (10) \qquad \widehat{S_p^{(t)}} = \alpha(t) \cdot \widehat{S_p^{(t-1)}} + (1 - \alpha(t)) \cdot S_p^{(t)} \quad (11)$$

The smoothing factor $\alpha(t)$ is as a function of trials, as shown in equation 12 where $\alpha_{\text{ceil}}$ sets the maximum amount of smoothing (usually 1.0) and $\lambda = \ln 2 / H$ determines how quickly smoothing ramps up and $H$ is the half-life (which can be set based on the need), meaning after $H$ trials, $\alpha(t)$ reaches 0.5. This formulation of $\alpha$ means that, early on (when there are few trials), scores are highly responsive to new data, while as the number of trials increases, the model becomes more stable and emphasizes longer-term performance trends.

$$\alpha(t) = \alpha_{\text{ceil}} \cdot \left(1 - e^{-\lambda(t-1)}\right) \quad (12)$$

It is to be noted that the leaf nodes in the hierarchy correspond to performance metrics (Level 4), which represent directly observable behaviors and are computed independently for each trial using the methods described in Sections 4.1 and 4.2. In contrast, higher-level nodes (Levels 0–3) capture latent constructs that are not directly observable and are aggregated after each trial using the procedure outlined above to assess evolving performance trends.

## 5. RESULTS

### 5.1. Experimental Setup

To evaluate our methodology, we analyzed performance data from four representative teams within the SUPRA dataset, labeling them as Team 1 through Team 4. For each team, we calculated performance metrics across three consecutive trials to enable direct comparisons in a consistent training environment. All metric parameters and configurable variables were carefully reviewed and refined with input from subject matter experts experienced in ECR drills. In the CTA hierarchy, we assigned equal weights to all child nodes under each parent node, meaning each child contributed equally to its parent's overall score (i.e., each child's weight was the reciprocal of the total number of children). We used equal weighting in this demonstration to illustrate our metric calculation and hierarchical aggregation process. Experts can customize the metric parameters and node weights to align with specific training objectives. To improve interpretability, we established thresholds for each score, classifying performance as above, at, or below expectations based on the values shown in Figure 7 (see following page). These thresholds are also adjustable, allowing modifications to better fit specific training scenarios and observed performance patterns.

### 5.2. Experimental Results and Key Observations

Figure 7 summarizes the results of this experiment. The heatmap on the left side of the figure shows Team 3's combined scores across three trials. Overall, most high-level scores improved over time, indicating that the team learned and performed better with each trial. This trend aligns with the lower-level metrics, which generally increased from Trial 1 to Trial 3. One exception was the "Total Room Coverage Time," which decreased due to the team's delay in covering a corner near the door—they only turned towards it as they exited the room. However, this had minimal impact overall, as improvements in eight other situational awareness metrics helped offset it. Although "Role Clarity" showed a slight increase, it remained weak due to consistently low scores in both the associated metrics, Identify and Capture POD, and POD Capture Time.

After aggregating the higher-level construct scores across all three trials for all teams, the scores following Trial 3 are shown in the heatmap on the right side of Figure 7. Several notable trends emerged. Role Clarity (Level 3) was significantly lower for Teams 1 and 3, mainly due to consistently weak performance in the Level 4 metrics Identify and Capture POD and POD Capture Time, observed across multiple trials, including the 3rd trial as shown. In contrast, Team 4 demonstrated consistently strong performance in these metrics, leading to higher aggregated scores for Role Clarity and highlighting the hierarchical influence of lower-level metrics. Team 2 had the lowest Cooperation (Level 2) score, driven by persistent poor performance in the Entrance Vectors and Teammate Coverage metrics, as seen even in the final trial. Notably, Team 4 consistently achieved high scores across all levels, reflecting cohesive team behavior and tactical effectiveness, supported by strong and stable performance across nearly all Level 4 metrics throughout the trials.

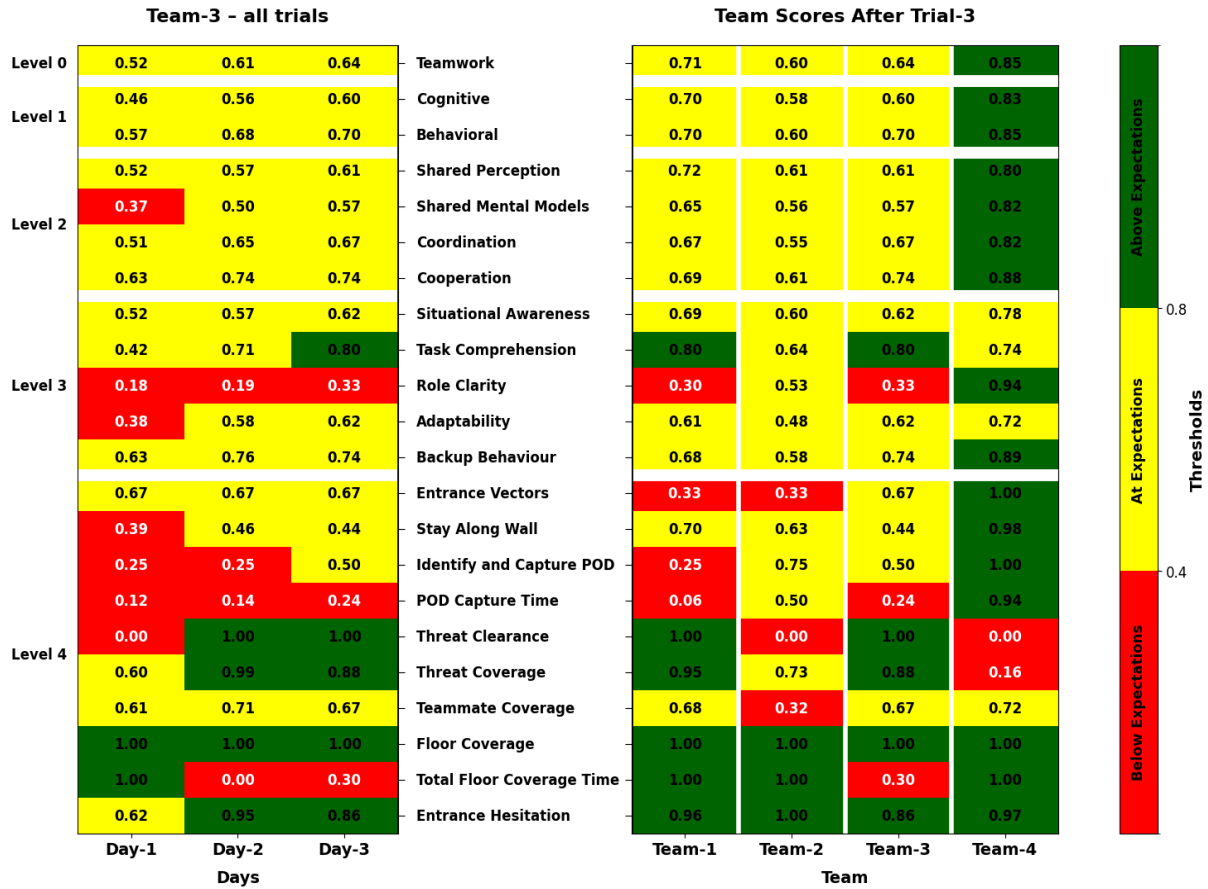

**Figure 7. CTA Hierarchy Performance Scores:** *(Left) Heatmap showing Team 3's scores across three consecutive trials; (Right) Heatmap displaying final aggregated scores for all four teams after completing three trials.*

In the hierarchical CTA structure, higher-level constructs such as "Teamwork," "Cognitive," and "Behavioral" clearly reflect the cumulative effect of lower-level metrics. Teams that consistently perform well across these metrics generally achieve higher overall scores, while specific weaknesses are also easy to identify. Overall, Team 4 ranks the highest, followed by Team 1, Team 3, and Team 2. This ranking aligns well with qualitative assessments from subject-matter experts who reviewed the respective drill sessions used in this study.

## 6. DISCUSSION AND FUTURE WORK

Based on the findings reported in this paper, we explore the broader applications, generalizability, and practical implications of our proposed methods. We also discuss the potential for supporting AARs and highlights current limitations and areas for future research.

The methods described in Section 4 show promise for broader use beyond the SUPRA scenario. Although the approach for extracting skeletal, positional, and gaze data from a video, as outlined in Section 4.1, remains consistent across contexts, the specific models, parameters, and calibrations need to be adjusted to fit the unique characteristics of video data in the new application domain. In addition, the domain-specific metrics developed for this case study, as detailed in Section 4.2, can be applied to other ECR scenarios—whether live (real personnel training on real systems and equipment) or virtual (real personnel training on simulated systems and equipment). However, the selection and setup of metric parameters should be guided by their relevance to the drill's standard operating procedures (SOPs) and specific objectives. Furthermore, the roll-up approach described in Section 4.3 can easily accommodate the addition of new metrics or the integration of alternative CTA models developed for different drills.

Additionally, our metric generation and assessment framework can be extended to support AAR processes. Our current prototyping leverages the Generalized Intelligent Framework for Tutoring (GIFT), an adaptive training framework, specifically its Game Master User Interface (Goldberg et al., 2020), which offers interactive dashboards valuable for both instructors conducting AARs and trainees reviewing performance insights. We are developing this framework to include not only the generated performance scores, but also explainable visualizations linked directly to videos recorded during the trial, which help clarify why particular scores were assigned. Figure 8 shows how the dashboard can be used to verify and explain gaze-based metrics. Comparative features, such as expert trajectory comparison in Figure 9, will also be integrated. These features enable the comparison between observed and expert behaviors.

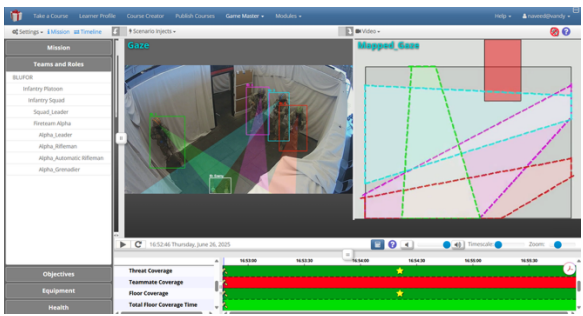

**Figure 8. AAR Dashboard** *(Showing gaze triangles generated on video and on the map, and below the videos are the metrics color-coded based on thresholding the scores)*

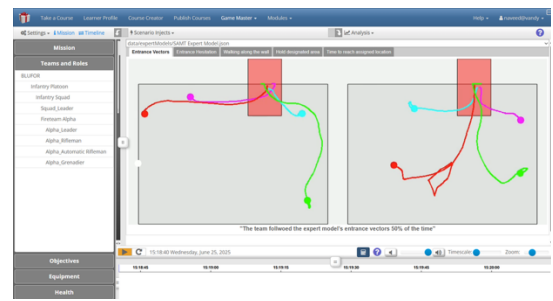

**Figure 9. AAR Dashboard** *(comparing expert trajectories (left) with the team (right))*

Despite promising results and the potential for broader uses of extracting skeletal, gaze, and position data from videos, several limitations still exist. First, while the tracking algorithm (OC-SORT) discussed in Section 4.1.2 performs consistently within our videos, Kalman-filter–based tracking methods may struggle to maintain identities in crowded scenes or when occlusions happen often. Second, the computer vision–based box and pose detection models used in Section 4.1.1 depend on factors like video quality, camera angles, and lighting, which influence detection accuracy and demand careful fine-tuning or retraining—sessions that can be resource-heavy. Third, the homography calibration described in Section 4.1.3 demands precise execution because mapping conversions are sensitive to small errors in selecting calibration points; any changes in camera positioning would necessitate recalibration to maintain accuracy. Fourth, regarding human evaluations, we only had access to overall qualitative performance scores provided by the Officer Commanding (OC). These scores capture broader subjective aspects in addition to what our metrics address and are not directly aligned with them. As a result, only the overall performance trend across selected teams could be compared, as discussed in Section 5.2. Finally, our current analysis has only used 2D pose estimation, which, although informative, naturally limits the richness of captured cues.

Future research will address these limitations through several targeted actions. Primarily, we aim to use fiducial markers to improve identity-tracking robustness. Fiducial markers, such as ArUco, AprilTag, or customized markers (Jurado-Rodríguez et al., 2021), provide reliable identification capabilities with minimal computational resources, making them particularly suitable for indoor scenarios like ECR drills. On the human evaluation side, we intend to design a standardized rubric aligned with our metrics and collect per-metric assessments from OCs, which can be compared to the metrics generated by our pipeline. This will allow for systematic comparisons between expert judgment and automated measurements. Additionally, we are currently working toward establishing a comparison of our video-based positioning approach using Mo-Cap data, the results of which will be reported in subsequent work. We also plan to collect sensor-based gaze information to enable direct comparisons with our video-based gaze approach. We will also explore video captured with depth cameras and develop methods for extracting 3D positional, gaze, and skeletal information to enrich behavioral interpretation. Lastly, we will conduct extensive cross-domain testing to validate the generalizability and applicability of our proposed methods across varied operational scenarios, and work toward extending the approach beyond room-scale drills to larger environments, enabling its application to company-level maneuvering and broader tactical exercises.

## 7 SUMMARY AND CONCLUSION

In this paper, we presented a comprehensive pipeline designed to maximize the utility of video data by extracting skeleton, position, and gaze information, thereby eliminating the need for intrusive, costly, and time-consuming sensor-based systems. Through a case study, we demonstrated the application of insights derived from this pipeline by introducing novel metrics alongside existing ones, focusing on positional dynamics, interaction patterns, and gaze behaviors within Enter and Clear Room scenarios. Additionally, we extended an existing Cognitive Task Analysis framework for modeling teamwork in these scenarios by integrating these newly developed metrics, thereby enhancing its capability. The results validate the practical usefulness and effectiveness of the information gained from video data using our pipeline. Furthermore, we discussed the generalizability of different aspects of our work and how the performance measures can be integrated into interactive, explainable dashboards to support after-action reviews and training management.

## 8 ACKNOWLEDGEMENTS

The research and development capabilities presented in this paper were directly supported by the US Army DEVCOM Soldier Center Cooperative Agreement (W912CG-22-2-0001). We want to thank Meghan O'Donovan, Clifford L. Hancock, and the entire SUPRA Team for their efforts in collecting and providing the data, as well as Staff Sergeant Giovanni O. Olivera Martinez for collaborating with us on the development and evaluation of the metrics. We are also grateful to John M. Roth, Ranges Team Lead Engineer at PM TRASYS, for his valuable feedback throughout several stages of the writing process, which greatly strengthened this paper. The views in this paper do not represent the position or policy of the United States Government, and no official endorsement should be inferred.